%% file: main.tex
\documentclass[krantz1,ChapterTOCs]{krantz} 

\usepackage{amssymb}
\usepackage{amsmath}
\usepackage{graphicx}
\usepackage{subfigure}
\usepackage{makeidx}
\usepackage{multicol}
\usepackage{algorithm}
\usepackage[noend]{algpseudocode}
\usepackage{url}
\makeindex
\algnewcommand\algorithmicforeach{\textbf{for each}}
\algdef{S}[FOR]{ForEach}[1]{\algorithmicforeach\ #1\ \algorithmicdo}
\begin{document}
\mainmatter
\setcounter{chapter}{4}
\include{ch5}
\bibliographystyle{unsrt}
\bibliography{bib.bib}
\printindex
\end{document}

%% file: ch5.tex
\chapter{Long-term area coverage and radio relay positioning using swarms of UAVs}
\chapterauthor {F. Benedetti} {Singular Perception s.r.l, benedetti.floriana@gmail.com}
\chapterauthor {A. Capitanelli} {Teseo s.r.l., alessio.capitanelli@teseotech.com}
\chapterauthor {F. Mastrogiovanni} {Universiy of Genoa, fulvio.mastrogiovanni@unige.it}
\chapterauthor {G. Vercelli} {University of Genoa, gianni.vercelli@unige.it}
\vspace{12pt}
\noindent{\fontsize{11}{11}\textbf{AUTHORS}}\normalsize
\vspace{5pt}
\\\textbf{Floriana Benedetti}, \textit{Singular Perception s.r.l.}, benedetti.floriana@gmail.com\\
\textbf{Alessio Capitanelli}, \textit{Teseo s.r.l.}, alessio.capitanelli@teseotech.com\\
\textbf{Fulvio Mastrogiovanni}, \textit{Universiy of Genoa}, fulvio.mastrogiovanni@unige.it\\
\textbf{Giasnni Vercelli}, \textit{University of Genoa}, gianni.vercelli@unige.it

\section{Introduction: Problem and Background}
\label{sec:introduction}
Unmanned Aerial Vehicles (UAVs) are becoming increasingly useful for tasks which require the acquisition of data over large areas. 
Despite that, many real-world scenarios still offer significant challenges due to communication issues and high costs.
The coverage problem, i.e., the problem of periodically visiting all subregions of an area possibly at a given frequency, is especially interesting because of its practical applications, such as surveillance, agricultural operations and continuous, long-term monitoring of areas hit by a natural disaster. 
We focus here on this last scenario, considering that in the case of events such as earthquakes, floodings or hurricanes, aerial coverage could help search and rescue teams to locate victims and assess the evolving state of the surrounding buildings and infrastructures. 

On the one hand, it is important to highlight the specific issues that set first response apart from other coverage and exploration applications, and why this is a challenging problem given the current state of the art. 
First of all, in this application the main interest lies in efficient long-term patrolling of a given area (\textit{Requirement} R1), whose map is supposed to be at least roughly available. 
This contrasts with the more popular exploration problem, where the focus is exactly on obtaining such a map as quickly as possible.
On the other hand, with respect to most commercial aerial coverage applications in the field of agriculture and surveillance, a hypothetical swarm of UAVs should be able to communicate back to the base station the gathered data in real time, without relying on pre-existing infrastructure (R2).
Furthermore, it should be robust to changes in the environment (R3) and to other technical difficulties, which naturally arise in long-term operations and cannot be reasonably foreseen before the mission starts, such as a UAV needing to recharge or experiencing a fault (R4). 

In such a scenario, R1 is implicitly solved by employing a coverage algorithm with one or more UAVs, while it is possible to satisfy R2 by adopting several supporting drones acting together as a signal relay chain to the base station. 
It is noteworthy that relay positioning has already been successfully employed to increase the effective communication range of mobile robots, route the signal around obstacles and drive down operation cost as smaller, simpler UAVs can be used rather than more expensive models. 
Despite this approach being common in exploration scenarios \cite {rooker2007multi} \cite {mukhija2010two}, to the best of our knowledge it has not been employed in coverage applications so far. 
Finally, R3 and R4 can be addressed using \textit{online} coverage algorithms able to dynamically react to a changing environment, instead of relying on \textit{offline} methods usually adopted in most coverage applications. 
Yet, a system of this sort can be defined only as a \textit{na\"{\i}ve} simultaneous coverage and relay positioning system, as it suffers from an evident problem, i.e., the growing number of UAVs which are not directly participating to the task at hand but are uniquely useful to support communication, something that effectively limits the overall system's efficiency.

In this Chapter, we propose an approach to greatly mitigate this issue by involving the supporting UAVs directly into the coverage process, in such a way that a drone acts in autonomy while the rest of the swarm provides a robust relay chain with the base camp and simultaneously help in the coverage operation, leading to \textit{real} simultaneous coverage and relay positioning.

Obviously enough, the conceptual framework described in this Chapter can be applied to any sort of mobile robots, but we will focus on the UAVs example given their suitability to face natural disasters, as they allow for quickly acquiring highly informative data from a given height, and are able to fly over areas potentially unreachable to ground robots or first responders in the case of a catastrophic event.

Throughout the Chapter, we will assume that a swarm of drones is tasked to repeatedly fly over an area just struck by a catastrophic event, in order to collect data about the environment and possibly locate survivors in need for help as they leave the buildings hit by the disaster. 
An indicative map of the area may be available, either from prior knowledge of the environment or acquired by the very same swarm in a preliminary exploration mission. 
The available map is then discretized into nodes with a suitable resolution, each linked to the reachable neighboring nodes through edges to form a connected graph. 
The available UAVs will travel from node to node, trying to keep the average number of node visits as homogeneous as possible, but also taking into account communication concerns. 
The system should be able to be robust to any of the UAVs going missing or being unable to operate, as a drone can: 
(i) lose connection with the swarm due to unforeseeable reasons, 
(ii) need to leave the swarm due to technical difficulties, either temporarily or permanently, (e.g., low battery, hardware faults), and 
(iii), be still part of the swarm but unable to leave the current node, as it is busy with a secondary task (e.g., offering assistance to or communicate with survivors, gathering specific data). 
In these pages we present an algorithm to deal with such events, we develop an architecture employing it and simulate its behavior in a Gazebo simulation with real UAVs' software in the loop. 
Finally, we compare our results to the case where a single drone is tasked to cover the area while the others only offer communication support (i.e., the na\"{\i}ve case). 
As we will see in the Conclusions, the proposed approach is able to provide a sensible improvement in coverage time compared to the reference method, with just a small hit on the average communication cost and barely affecting the worst-case communication cost. We conclude that such system can be deployed to swarms of UAVs already operating in the aforementioned mode in order to improve its performance.

\section{State of the Art}
\label{sec:sota}
As discussed above, the main idea underlying this Chapter is to jointly treat two distinct problems in the literature, namely \textit{area coverage} and \textit{relay positioning}. 
While both problems have been widely faced in the literature singularly, no technique has been formalized to achieve a functional trade-off between the two, similarly to what has been previously done with simultaneous exploration and relay positioning \cite{cesare2015multi}.
In this Section we will discuss the current advancements in both fields and present tools that can be used to address the issue at hand.

\subsection{Coverage}
\label{sec:coverage}
The problem of coverage is defined as the problem of sweeping a given area with the highest possible efficiency, i.e., being able to visit each point in that area with a possibly constant frequency that is as high and as homogeneous among all the nodes as possible. 
It is clear that this is very different from the more common shortest-path problem, where one is mainly interested in finding the minimum cost trajectory between any two points. 
Despite that, coverage also has a large number of important applications, ranging from harvesting \cite{couto2016model} to mine hunting \cite{nicoud1995pemex}, and this is why a multitude of algorithms have been proposed in the past. 
Historically, these algorithms are categorized in two ways: \textit{heuristic} vs. \textit{complete} \cite{galceran2013survey} algorithms, and \textit{online} vs. \textit{offline} \cite {choset2001coverage} algorithms. 
According to such classification, we define as complete only the algorithms which can be mathematically proven to obtain complete space coverage \cite {Khan2017OnCC}, while we define as heuristic the others, as in \cite {Alinaghian2015two}.
The second classification instead is more complex than it appears at first sight, due to its practical implications in certain applications:
\begin{itemize}
    \item \textit{Offline methods} precompute a (often optimal) coverage strategy for a given area before runtime operations occur, using static \textit{a priori} information. Notable examples are Space Decomposition \cite{choset2000coverage} and Spanning Tree Coverage \cite{jones2009cooperative}.
    \item \textit{Online methods} do not perform calculations offline based on \textit{a priori} knowledge, instead they periodically determine at runtime the next best action to achieve coverage based on current environmental knowledge and sensor input. The most famous example is given by Real Time Search \cite {cabreira2016terr}.
\end{itemize}
Historically, offline methods have been the preferred choice in most cases, as the majority of the coverage applications happens to deal with mostly static environments, for example large fields in either agricultural or security applications \cite{valente2011waypoint}, as well as with very structured environments, like perfectly known surfaces in painting applications \cite {vempati2018paintcopter}. 
In the case presented here though, online methods are the preferred choice as they overcome some limitations of offline methods, which are clearly in contrast with the requirements of the proposed scenario:
\begin{itemize}
    \item it is not possible to assume perfect \textit{a priori} knowledge of the environment in the event of a disaster;
    \item it is computationally expensive to precompute or re-compute a strategy for large areas under optimality constraints;
    \item offline algorithms exhibit very low tolerance to unpredictable event, as they cannot be taken into account when the strategy is computed offline.
\end{itemize}
As a notable consequence to the last point, offline methods strongly depend on every single UAV part of the swarm, as they assign to each UAV a list of nodes to visit.
If that UAV is busy, missing or otherwise unable to act, that would cause a major failure of the system.

Online coverage algorithms obviously pose a greater challenge to ensure efficient operations with multiple UAVs compared to offline ones. 
As an example, it is often necessary to make a number of assumptions about which knowledge can be shared among UAVs, something that is not necessary if the whole strategy is computed offline. 
Examples of algorithms employed in the online multi-agent case are Node Count \cite{korf1990real}, LRTA* \cite{korf1990real}, Edge Counting \cite{koenig1996easy} and PatrolGRAPH* \cite{baglietto2008patrolgraph}.
While it is not surprising to know that increasing the number of UAVs generally leads to better coverage performance, it should be noted that more complex algorithms that may expand the single agent case, can be overtook by simpler algorithms in the multi-agent case \cite{nattero2014coverage}. 
As an example, Node Count \cite{balch1993navig} is the simplest algorithm among the cited ones, but also the one that performs better in comparative benchmarks \cite{nattero2014coverage}. 
It fundamentally consists in maintaining a database of the number of times each node has been visited, share it among the agents, and let each agent move toward the neighboring reachable node with the lowest score. 
In the following Sections we show how this idea can be used to achieve simultaneous relay positioning and area coverage.

\subsection{Relay Positioning}
\label{sec:rp}
Optimal relay positioning is a widely explored problem in the telecommunication field, where it is used to place antennas in a given area to provide robust wireless connection \cite{shen2009heuristic}\cite{younis2008strategies}. With mobile robots though, a new problem arises, that is how we can create robust \textit{dynamical} relay chains able to rearrange themselves online \cite{chen2017optimal}\cite{jiang2010dynamic}.
Keeping in mind that it is generally desirable to relay a signal from one point to another, this translates in our case into relaying data from one far UAV back to the base station, through a number of intermediate UAVs. 
This implies that only the leading UAV, the master, benefits of complete freedom of action, while the other UAVs, which must primarily provide communication support, are forced to assume the positions imposed to them by the relaying algorithm.
Furthermore, it is noteworthy that this kind of algorithms not only provides a viable relay chain, but also strive to achieve one of the following secondary goals, or a balance of the two: 
(i) maximizing the quality of the resulting chain in terms of inter-agents and master-base communication, and 
(ii) minimizing resource requirements, i.e., the number of agents employed in the chain \cite{burdakov2010relay}\cite{olsson2011pos}.
In Section \ref{sec:sota} we will detail how one can dynamically generate chains of optimal communication cost with the available resources, and how this can be relaxed to guarantee supporting UAVs with a certain degree of movement margin in the coverage process.

\section{Methodology}
\label{sec:methodology}
In this Section we show how it is possible to combine relay positioning and coverage algorithms. 
Such an integration will be achieved by letting a single UAV, i.e., the master, act freely, either directly controlled by an operator or automatically by a coverage algorithm, while the rest of the swarm offers communication support in the form of a relay chain. 
The main difference with previous approaches in the literature is that we relax the communication optimality criteria of the chain to allow the supporting drones to increase the rate at which the area is covered. 
To do so, we first formalize the problem of dynamical single-target relay positioning, then we proceed by illustrating a typical solution in the case in which one wants to achieve the best optimal chain given a limited number of UAVs. 
We conclude the Section introducing a new cost term penalizing support drones entering areas which have already been covered multiple times, effectively improving the efficiency of the coverage task.

\subsection{Dynamical Single-Target Relay Positioning}
\label{sec:dstrp}
Let us provide a more formal description of the dynamical single-target relay positioning problem as described in \cite{burdakov2010relay}. 
Let $x_0, x_n \in \mathcal U$, where $x_0$ is the position of the base station, $x_n$ is the position of the UAV we want to relay to the base, and $\mathcal U$ is the space (bi-dimensional or tri-dimensional) of positions achievable by the UAVs. 
Given any two points $x, x'$ in $\mathcal U$, we define the \textit{communication reachability function} $f_{comm}(x, x')$ as a function that yields $1$ if the communication between the two points is feasible, and $0$ otherwise. 
It is now possible to define a relay chain $rc_j$ as an ordered set $rc_j = (x_0, x_1, \ldots , x_n)$, such that $x_0, x_1, \ldots, x_n \in \mathcal U$ and $n$ is the \textit{length} of the chain. 
Such chain is said to be \textit{valid} if and only if $f_{comm}(x_i, x_{i+1}) = 1$ for each $i = 1, \ldots, n-1$. 
Finally, let us introduce the \textit{communication cost function} $c_{comm}(x, x')$, which models the cost of transmitting information from an agent placed in $x$ to another one placed in $x'$. 
The total communication cost $C_{comm}$ of a given chain can then be given by the relation:
\begin{equation}
    \label{eq:cost}
    C_{comm}(rc_j) = \sum^{n - 1}_{i = 0} c_{comm}(x_i, x_{i+1}). 
\end{equation}
Additional costs may be taken into account to model specific desired properties of the chain (e.g., the distance of the last element of the chain from a specific target), otherwise one can assume the total communication cost as equivalent to the total cost of the chain $C_{tot}$. 
It is noteworthy that we will not give a specific definition for $c_{comm}(x, x')$, as this cost depends on the actual implementation and the way one desires to model the quality of data transmission between any two points (e.g., strength of the signal, number of packets lost).

Once the definition of total cost of the chain has been given, one can generate different kinds of chains between any two points by solving different sub-problems, such as finding the chain of minimum length with minimum cost, or conversely, finding the chain of minimum cost with minimum length. 
In most practical applications with UAVs though, the focus is on finding a minimal cost chain that employs at most the number of UAVs available for the specific application at hand.

\subsection{Determination of the Limited Length Chain of Minimal Cost}
\label{sec:mlmc}
In this Section we present how to compute the chain of minimal cost given a limited number of UAVs available for the relay chain based on the methods described in \cite {burdakov2010relay}, following the definition of the \textit{hop-constrained shortest path problem} \cite{dahl2004directed}. 
This approach is based on a combination of the \textit{Dual Ascent method} \cite{balakrishnan1992using} and the \textit{Dijkstra's algorithm} \cite {burdakov2010relay}. 
Please note that in our scenario the Dijkstra's algorithm is not used to generate a path, but a \textit{chain}, i.e., the resulting series of nodes is not meant at being traversed by a single UAV but it is rather a list of positions which must be occupied by the available UAVs to form a valid relay chain. 
The basic idea is to use the Djikstra's theorem to get a solution as a series of nodes from the environment map going from the base station to the last, independent, UAV of the chain, without regard for the resulting series' length. 
If the solution initially obtained satisfies the limit on the number of UAVs, it is immediately accepted; otherwise the Dual Ascent algorithm is used to refine the costs used by the Djikstra's algorithm in a way to favor shorter length solutions. 
The two algorithms are then iterated until a valid solution is found or no further refinement is possible. 
In particular, our use of the dual ascent algorithm is detailed in Algorithm \ref{1}.

More formally, let $N$ be the set of all nodes $n$ which the agents can occupy, each one corresponding to a given position $x$ in the map, and let $n_0$ be the node corresponding to the first fixed element of the chain (i.e., the base station). 
Furthermore, any of such nodes can be connected to several peers by directed edges $(n, n')$, which form the set $E$. 
Together, $N$ and $E$ define the navigation graph $G$, i.e., a connected, non-oriented graph of arbitrary order, which may include cycles. 
Considering that each node $n$ is associated with a position $x$, the cost definitions given in the previous Section hold equally for both positions and nodes, as long as the cost of a node $n$ is computed on its assigned position $x$. 
In particular, we remark that $c_{comm}(n, n') \equiv c_{comm}(x, x')$, allowing us to compute the tree of minimum length minimum cost chains (MLMC) on the navigation graph $G$, that is the tree of the optimal chains in terms of cost and length (i.e., the number of employed UAVs), from $n_0$ to all the other nodes in $N$. 
This can be achieved by using the version of Djikstra's algorithm, which is modified in such a way that: 
(i) given a chain $\pi$, the cost of a chain is actually a compound of the form $(len(\pi), cost(\pi))$, and a chain ordering is induced as $(len_1, cost_1) < (len_2, cost_2)$ if and only if $(cost_1 < cost_2) \lor (cost_1 = cost_2 \land len_1 < len_2)$, i.e., cost has higher priority than length; 
(ii) the algorithm returns the whole tree it has built, instead of just the minimum cost chain between any two given nodes. 

\begin{algorithm}
\caption{Dual Ascent algorithm}
\label{1}
\begin{algorithmic}[1]
\State $\alpha \leftarrow \alpha_0$
\Loop 
\State Calculate MLMC tree from $n_0$ using $c'_{comm}(n, n') = c_{comm}(n, n') + \alpha$
\State Obtain from MLMC $y_n$, $q_n$ $\forall n \in N$ and $\pi$ from $n_0$ to $n_t$
\If{$len(\pi) \leq n_{UAVs} + 1$} 
\State{\textbf{return} $\pi$}
\EndIf
\State $S \leftarrow \{(n, n') \in E: q_{n'} > q_n + 1\}$
\If{$S = \emptyset$}
\State{\textbf{fail}}
\EndIf
\State $\epsilon_{n, n'} \leftarrow \frac{(y_n + c_{comm}'(n, n')) - y_{n'}}{q_{n'} - (q_n + 1)} \quad \forall (n, n') \in S$
\State $\epsilon \leftarrow$ min $\epsilon_{n, n'}$
\State $\alpha \leftarrow \alpha + \epsilon$
\EndLoop
\end{algorithmic}
\end{algorithm}

Before proceeding with the Dual Ascent algorithm, let us define a few more important quantities: 
(i) $n_{UAV}$, being the maximum number of UAVs available to build a valid chain; 
(ii) $q_n(n)$, being the depth of a node $n$, i.e., its distance from $n_0$ in terms of hops in the MLMC tree; and 
(iii) $y_n(n)$, being the current \textit{modified cost} of the chain from $n_0$ to $n$, that is the cost of the chain obtained by adding to the cost of each edge in the chain an additional component $\alpha$, usually initialized to $0$ and updated by the Dual Ascent algorithm at every iteration to increasingly favor shorter paths in terms of hops. 
Let us now have a look at Algorithm \ref{1}, which starts by computing the MLMC tree (line 3) with each edge cost increased by $\alpha$.
It is noteworthy that this implies that also $q_n(n)$ and $y_n(n)$ for every $n$ in $N$ are known after this step (line 3). 
If the MLMC tree contains a chain $\pi$ between $n_0$ and the node currently occupied by the chain's master $n_t$, using at most $n_{UAV}$ drones (line 4), then a valid solution has been found (line 5); 
otherwise, the MLMC tree may be refined to favor shorter solutions. 
This is achieved by computing $\epsilon$, which is the amount by which $\alpha$ should be increased to ensure that next time the Djikstra's algorithm will be called, the resulting tree will yield at least one shorter chain (lines 7-11). 
It is important though that the choice of $\epsilon$ ensures that no solution can be missed. 
To this purpose, given any node $n'$, let us consider all edges $(n, n') \in E$ which are not part of the current MLMC tree. 
One of such edges constitutes a strict improvement if added to the MLMC tree, if and only if the relation $q_n' > q_n + 1$ holds; we call the set of such edges $S$. 
Hence, by computing the value $\epsilon_{n, n'}$ by which all edge costs should be increased for an edge $n'$ in $S$ to be included in the MLMC tree, and taking the minimum among all those computed, we obtain an $\epsilon$ with the desired characteristics.
Following from such statements and without delving into the mathematical details, the value $\epsilon_{n, n'}$ can be obtained by applying the following equation:
\begin{equation}
    \label{eq:eps1}
    \epsilon_{n, n'} = \frac{(y_n + c_{comm}'(n, n')) - y_{n'}}{q_{n'} - (q_n + 1)} \quad \forall (n, n') \in S,
\end{equation}
where $c_{comm}(n, n')$ is the communication cost between two nodes, and $c_{comm}'(n, n')$ is the current modified cost, such that:
\begin{equation}
    \label{eq:eps2}
    c_{comm}'(n, n') = c_{comm}(n, n') + \alpha.
\end{equation}

Finally, if $S$ is empty, then no chain can possibly be shortened and Algorithm \ref{2} fails (line 7).
It is noteworthy that while the Algorithm may seem computationally complex, it is actually possible to optimize the process, for example by repairing the MLMC tree instead of computing a new one from scratch every time, as described in \cite{burdakov2010relay}. 

\subsection {Reformulation of the Chain Cost Function}
\label{sec:chaincost}
We now want to modify the cost function defined in Equation \ref{eq:cost} to favour nodes that have been visited fewer times, by introducing a term for the \textit{coverage cost} between two nodes $c_{cov}(n, n')$, which will behave similarly to the node count algorithm. 
To that aim, let us highlight the differences between \textit{Node Count} and \textit{Edge Count}: 
on the one hand, the former counts the number of times a node has been occupied at each iteration, and an UAV would move from the current node to the neighbouring connected node that has been previously visited less; 
on the other hand, the latter counts the number of times any edge has been traversed and strives to make sure that all edges of a given node are traversed with the same relative frequency. 
While node count gives the best result in the multi-agent case, edge count can be more naturally integrated with the Djikstra's algorithm and the definition of $c_{comm}(x, x')$ given in Section \ref{sec:dstrp}, as it assigns costs to edges, not nodes. 
Hence, we can slightly modify the common structure of node count by working in terms of edges like in edge count, and assigning to all inbound edges to a node the same count value, i.e., the number of times the destination node has been visited. 
More formally, from the standard Node Count procedure in Algorithm \ref{2}, we define an operator $count(n_i)$ that measures the amount of visits to a node. 
At each instant, the agent will select the less visited node from those in the \textit{neighborhood} of its current location $n_a$, i.e., the set of all nodes $n_i$ for which there exists an edge $(n_a, n_i)$. 
We expand this basic definition stating that there exists a similar operator for edges $count(n_a, n_i)$, such that $count(n_a, n_i) = count(n_i)$. 
Hence, the agent will select the edge leading to the less visited node in its neighborhood.
From this definition, it follows that edges must be directed, meaning that the edge $(n, n') \ne (n', n)$, and consequently the associated costs may be different as well. This may not be evident while dealing with communication costs only, as communicating from one point to another is just as difficult as the opposite, and hence the costs is the same, but it does matter to the purpose of computing the coverage cost. 

\begin{algorithm}
\caption{Node Count algorithm}
\label{2}
\begin{algorithmic}[1]
\Require The connected graph $G$, and an agent's current location node $n_a$
\Loop 
\State $n_a \leftarrow oneof(arg min_{n_i \in neighborood(G, n_a)} count(n_i)$
\State $count(n_i) = count(n_i) + 1$
\EndLoop
\end{algorithmic}
\end{algorithm}

We can now provide a definition for the term $c_{cov}(n, n')$ representing the coverage cost between any two nodes $n$ and $n'$. 
To this purpose, multiple candidates are equally legitimate as long as the term is directly proportional to the count associated with the edge $(n, n')$. 
Following the discussion above, the simplest candidate is then given by:
\begin{equation}
    \label{simplecost}
    c_{cov}(n, n') = \beta count(n, n'),    
\end{equation}
where $\beta$ is a scaling factor whose value is determined based on the order of magnitude of the communication cost, and how much the designer wants to favor coverage performance over communication stability and vice-versa.
It follows that the total cost of an edge becomes:
\begin{equation}
    c_{tot}(n, n') =  c_{comm}(n, n') + c_{cov}(n, n'),    
\end{equation}
while the total cost of a valid chain becomes:
\begin{equation}
    \label{eq:finalcost}
    C_{tot}([n_0, \ldots, n_t]) = \sum^{t - 1}_{i = 0} c_{tot}(n_i, n_{i+1}).
\end{equation}

In this regard, the reader may notice that, in long-term operation contexts, the coverage component may grow unbounded, leading to chains which increasingly favor coverage over communication. 
It is possible to address this issue in several ways, for example by normalizing the count value for every edge across the graph, and/or computing $c_{cov}(n, n')$ over a sliding time window.

In conclusion, by employing the cost in Equation \ref{eq:finalcost} to compute a modified MLMC tree and by substituting $c_{tot}$ to $c_{comm}$ in Equations \ref{eq:eps1} and \ref{eq:eps2}, we obtain a solution to the relay positioning problem which also takes into account the necessity to fly over under-visited nodes. 
No changes to the Algorithms \ref{1} and \ref{2} should be made to ensure the correct operation of the system.

\section{System's Architecture}
\label{sec:architecture}
\begin{figure}
  \centering
  \includegraphics[width=1\textwidth]{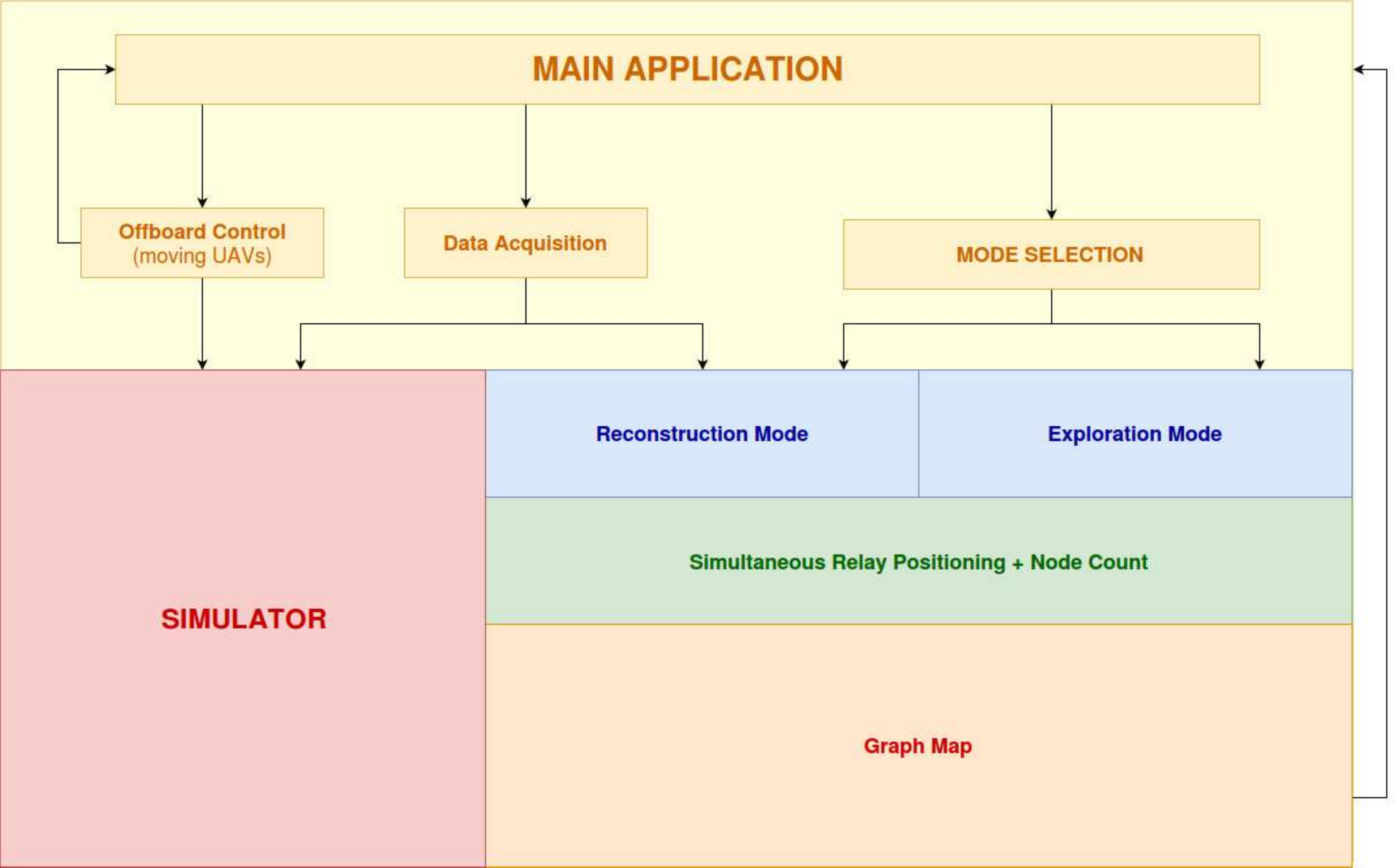}
  \caption[The proposed architecture.]{The proposed architecture. The main application acquires the UAVs’ positions and checks whether a new target has been found. If this is the case, it changes the master UAV to the one closest to the target. Then, after determining the new master position, computes the hybrid relay chain and dispatches the target positions to all available UAVs.}
  \label{fig:arch}
\end{figure}

Following the methods presented in the previous Section, we will now proceed to illustrate a sample architecture that can implement the ideal system outlined in Section \ref{sec:introduction}. 
Figure \ref{fig:arch} depicts the architecture structure and its main components. 
The main idea of the architecture is to act in \textit{interleaved rounds}, which means that initially the master UAV acts, or at least takes a decision on its next action, then the swarm reacts, by computing a valid hybrid relay-coverage chain and taking the corresponding configuration. 
Once all the UAVs have reached their designed destinations, the process starts over. 
It is noteworthy that the master UAV does not necessarily have to follow a coverage algorithm as long as the nodes it visits are recorded together with the ones visited by the rest of the swarm. 
Obviously enough, letting the master UAV act on pure \textit{Node Count} and share the map with the swarm's simultaneous relay chain and coverage algorithm is the most natural solution, and the one we explore here, but also manual control by an operator is a viable alternative.

\subsection{Communication Cost and Map Discretization}
\label{sec:cost}
We define $c_{comm}(n, n')$ as a positive monotonic function directly proportional to the shortest distance between the nodes $n$ and $n'$, and ranging from $0$ in the ideal case where the two nodes coincide, and $c_{comm_{max}}$ when the distance between the two nodes is greater than a given threshold distance $d_{comm_{max}}$. 
Similarly, we define $f_{comm}(n, n') = 0$ if the distance between the two nodes is greater than $d_{comm_{max}}$, and $1$ otherwise. 
Then, we arbitrarily discretize the map in a finite number of nodes, and define edges between all pairs of nodes for which $f_{comm}(n, n') = 1$ holds true. 
In this way, we ensure that when we generate a chain, communication is always possible between any two nodes connected by an edge while also providing a high number of connections, hence obtaining a valid discretized map rich of alternative paths where to compute MLMC trees. 
Finally, edges traversing large obstacles are pruned from the tree, either systematically or as a direct consequence of $c_{comm}(n, n')$, if the communication cost is already taking obstacles into account.

\subsection{Robustness to Secondary Tasks and Missing UAVs}
\label{sec:tasks}
As described in Section \ref{sec:introduction}, one or more UAVs may be busy with a secondary task, which requires them to fly over a given node for a prolonged amount of time. 
A robust solution to this issue as long as at most one UAV is involved in other tasks at each time instant, is to promote such UAV to master, and let the rest of the swarm rearrange accordingly to the simultaneous coverage and relay positioning algorithm. 
Since the master UAV is not required to have any specific capability, i.e., the distinction from the rest of the swarm is only logical, the master UAV can be changed on the fly without consequences. 
To reinforce this point, it is noteworthy that what we get from the algorithm is just a chain of nodes without any reference to a specific UAV. 
Clearly, while the new master UAV is busy in the same node over several rounds, the rest of the swarm will still be providing assistance with the coverage task and update the node counts accordingly. Once the secondary task is complete, the master UAV privileges can be returned to the original UAV or the execution can proceed with the new one. 
If more than one agent is expected to be busy with a secondary task at any time, then a more complex solution is necessary, e.g., implementing a queue and/or modifying the tree generation algorithm to enforce solutions including the nodes corresponding to the desired tasks' positions.

Another issue raised in Section \ref{sec:introduction} is how to provide robustness to the absence of any single UAV, either when this is planned, e.g., in the case of UAVs requiring a recharge, or an unexpected event, e.g., when an UAV is gone missing. 
In this case, a distinction should be made between the master UAV and rest of the swarm. If the master UAV leaves the swarm, either suddenly or in a regulated manner, there would be only minor consequence, as the chain would not be damaged and the only action required would be selecting a new master UAV and reducing the number of available UAVs for the simultaneous coverage and relay positioning algorithm (unless a substitute is readily available). 
On the contrary, the same does not apply for members of the relay chain. 
If the UAV is scheduled to leave the chain, preliminary measures can be taken before removing it and reducing the number of UAVs available for the algorithm. Otherwise, the only way to ensure that a chain would remain valid even if one or more UAVs were missing, is to use an overly restrictive communication cost policy in order to ensure (or at leastpromote) two non-consecutive UAVs to still be able to communicate. 

Any missing UAVs returning to the swarm can be easily reintegrated by increasing again the number of agents available to the algorithm. 
UAVs which got detached from the swarm have not access to the updated number of visits for each node but, assuming they have a reasonable amount of computing power onboard, they can still fall back to pure node count. 
As they approach less visited areas of the map, they have a chance to join again the swarm and synchronize their respective knowledge about visited nodes. 
If this is not possible or the UAV cannot regroup with the swarm, it would still be able to fly back to the base station.

\section{Experimental Results}
\label{sec:results}
\begin{figure}
  \centering
  \includegraphics[width=1\textwidth]{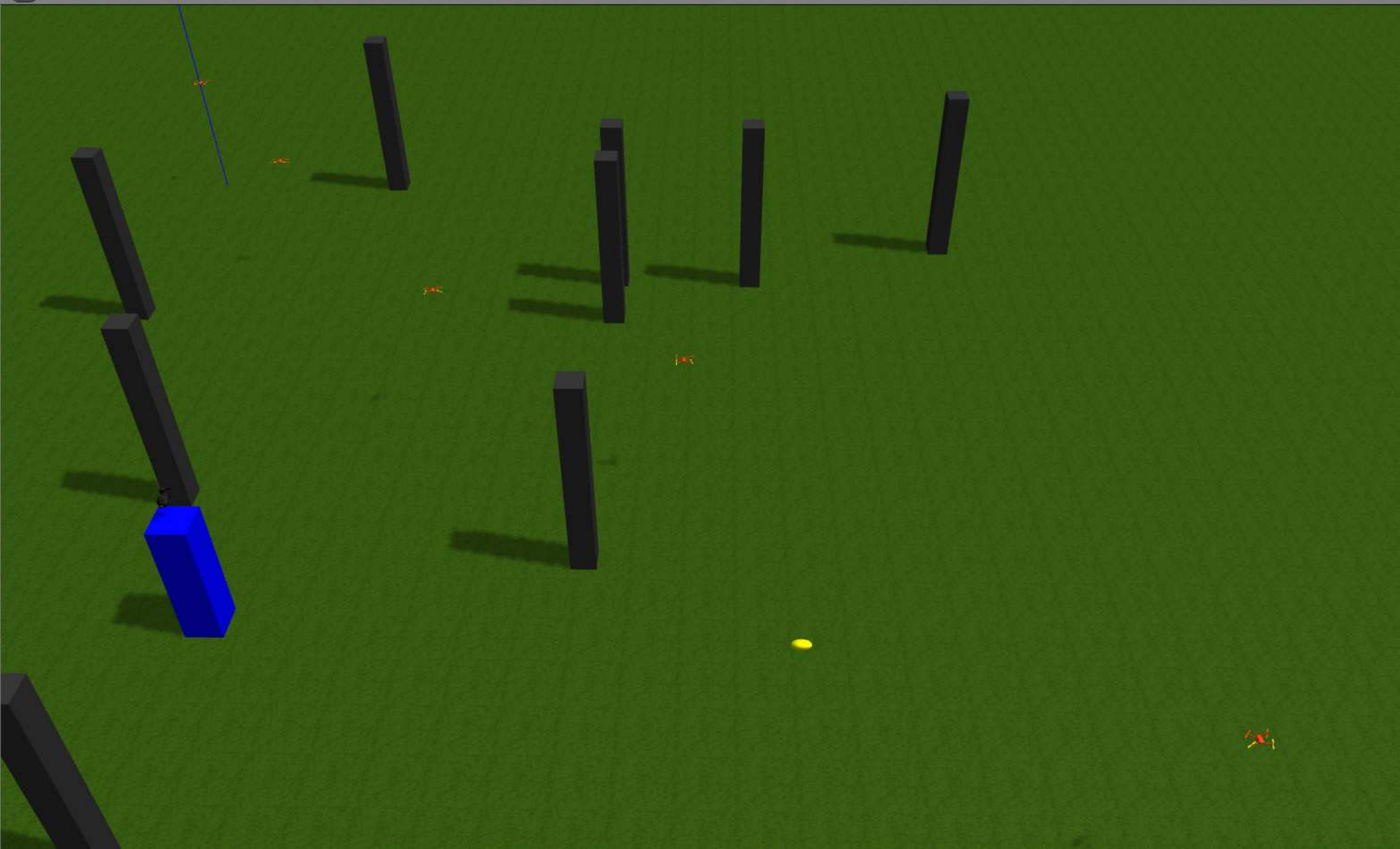}
  \caption[An example of the performed simulations.]{An example of the performed simulations. Brown boxes corresponds to obstacles, while the blue one is a target.}
  \label{fig:2}
\end{figure}

In this Section, we present the experiments used to assess the performance of the simultaneous coverage and relay-positioning algorithm and the related architecture, as presented in Section \ref{sec:methodology} and Section \ref{sec:architecture}, respectively. 
Experiments have been run in simulation as we are not interested in the dynamics of the UAVs in this context. 
However, we run UAV software in the loop to make sure the proposed algorithm is compatible with real-world UAVs. 
The UAVs we simulated are 3DR IRIS \cite{iris}, while the simulation environment is Gazebo \cite{gazebo}, running on a machine with Intel I7 4500U@3.00GHz processor and $8$ GB of RAM. 
The proposed architecture has been implemented in the Robot Operating System (ROS) \cite{ros} as a single module acting as a base station and sending commands to multiple instances of the UAV software, which run in parallel to simulate both the master UAV and the fellow UAVs available for the hybrid relay-coverage chain. 
Communication between the base station and the simulated UAVs is implemented using MAVLink \cite {mavlink} through the MAVROS bridge \cite {mavros}. 
A total of 5 UAVs have been simulated in a squared environment measuring $70 \times 70$ m, discretised in 196 nodes, each $5$ m apart from the others. 
Several tests were run with these settings with a number of randomly positioned obstacles and targets. 
When an UAV enters a node containing a target, a secondary task as the one described in Section 5.4.2 begins. 
In this particular case, the UAV flies around the target and gather point cloud data until a complete 3D model of the target is acquired. 
A sample simulation scenario is reported in Figure \ref{fig:2}.

All the tests have been conducted using a formulation for $c_{comm}(n, n')$ which increases with the distance between $n$ and $n'$, plus a penalty for the surrounding volume occupied by obstacles to model communication difficulties in cluttered environments. 
For $c_{cov}(n, n')$ we employed the simple definition given in Equation \ref{simplecost} and repeated the test several times for different values of the parameter $\beta$. 
Two kinds of tests were run, one where all nodes are to be visited at least once and one where they must be visited at least $10$ times. 
In all tests the following metrics have been measured: 
(i) the number of iterations needed to achieve the results, 
(ii) the number of times each node has been visited, 
(iii) minimum, maximum and average communication costs among all generated chains. 

\begin{figure}
  \centering
  \includegraphics[width=1\textwidth]{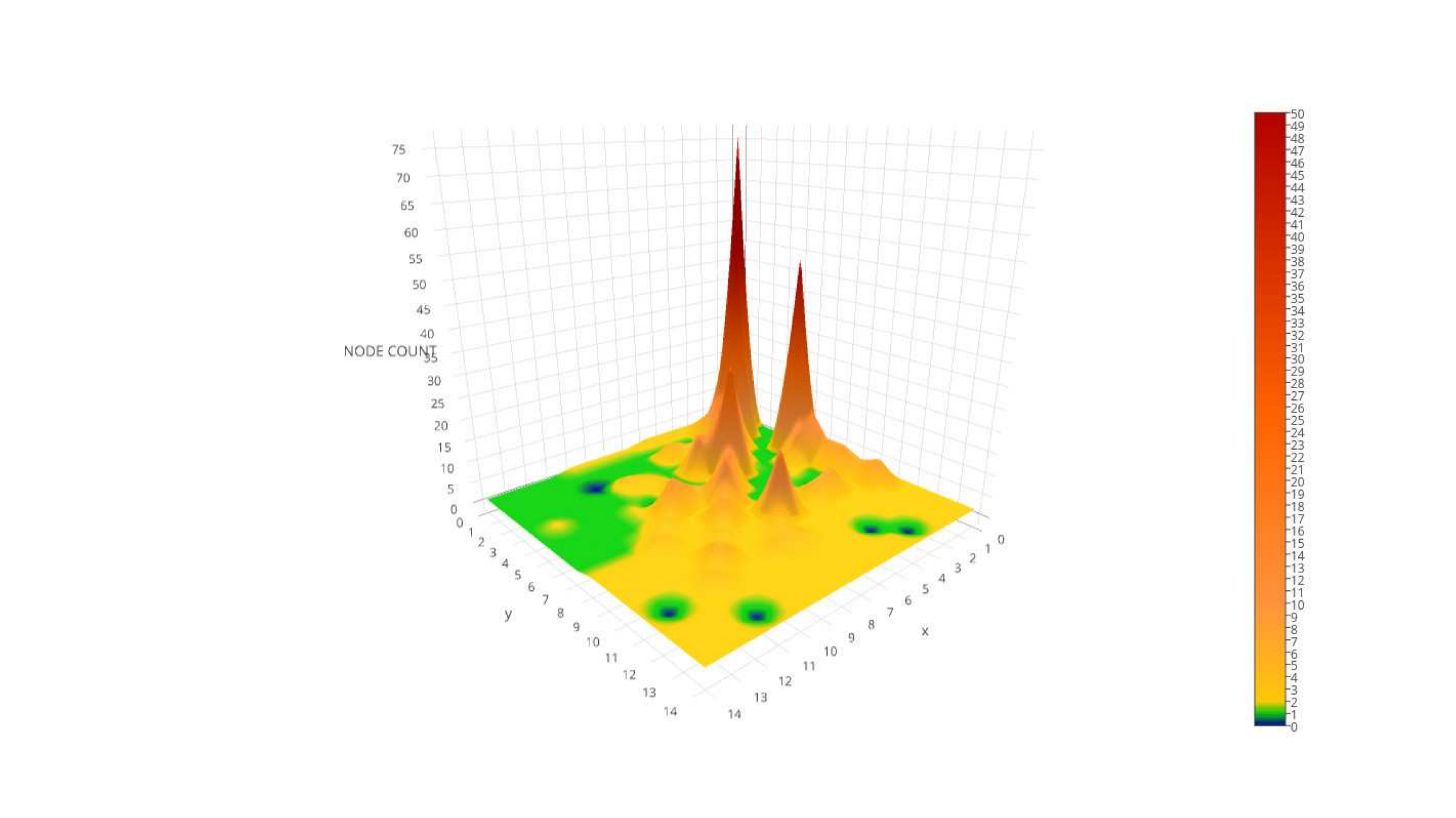}
  \caption{Number of visits for each node after they have all been visited at least once for $\beta$=0.}
  \label{fig:3}
\end{figure}

\begin{figure}
  \centering
  \includegraphics[width=1\textwidth]{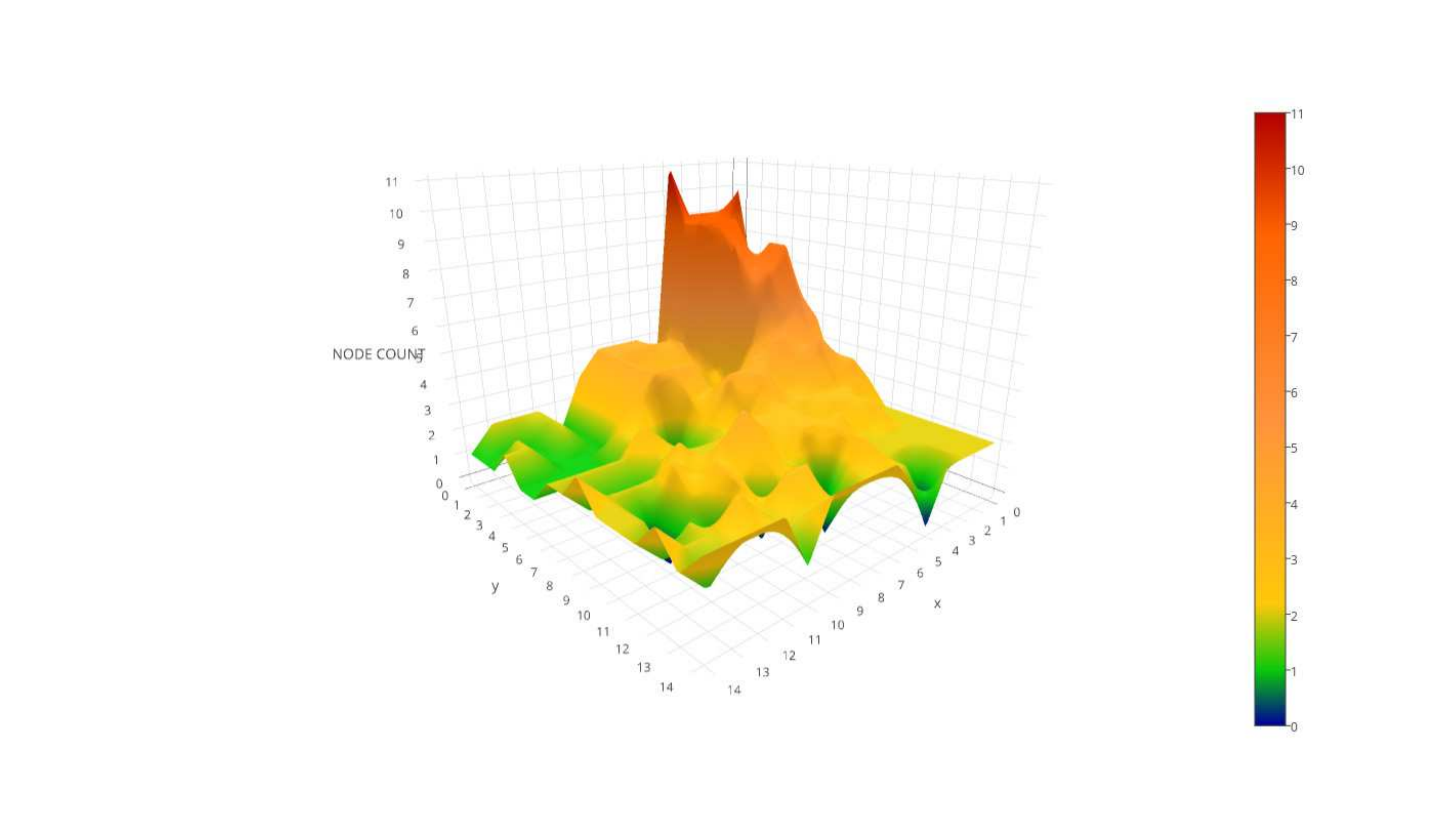}
  \caption{Number of visits for each node after they have all been visited at least once for $\beta$=0.5.}
  \label{fig:4}
\end{figure}

\begin{figure}
  \centering
  \includegraphics[width=1\textwidth]{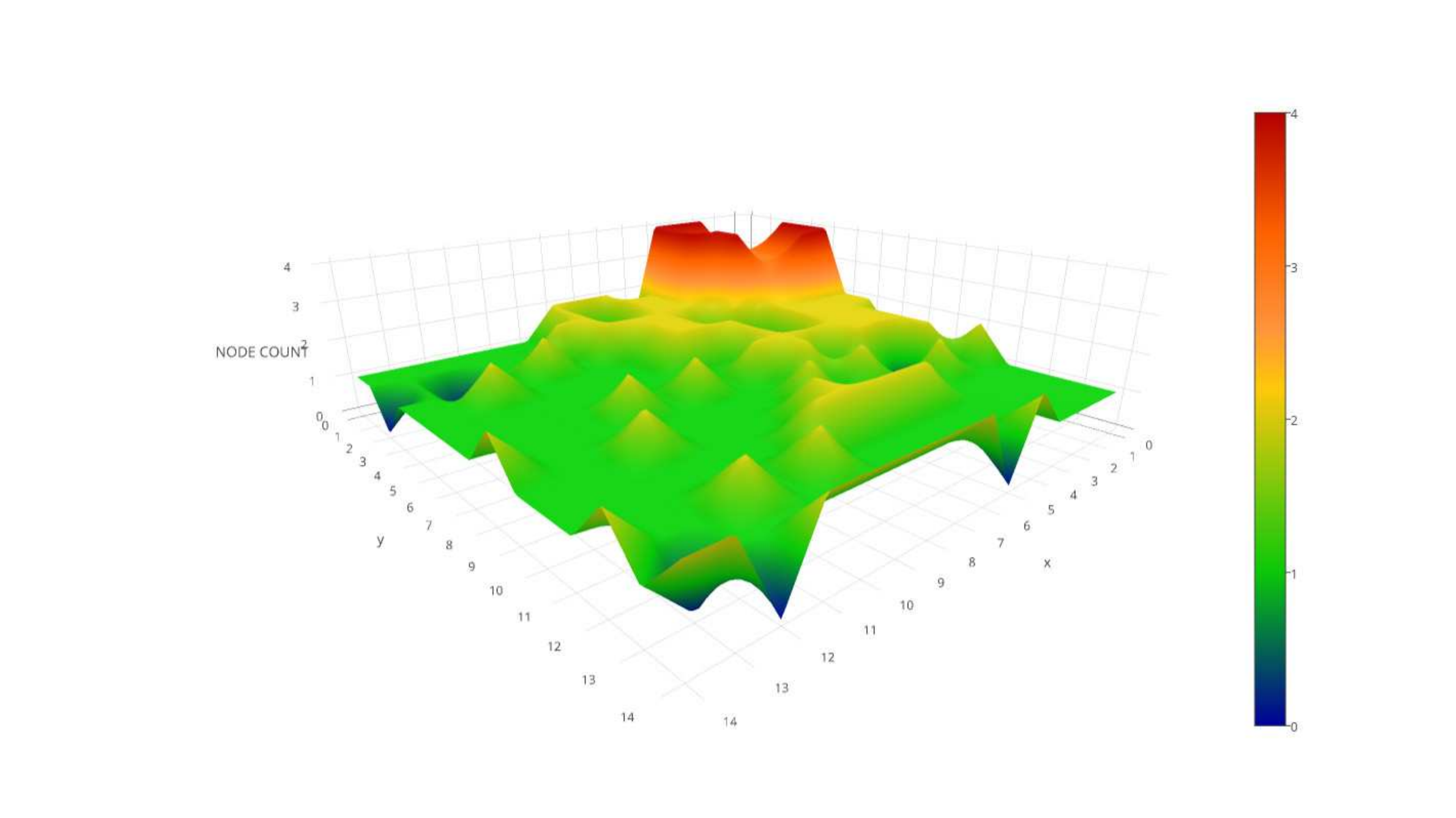}
  \caption{Number of visits for each node after they have all been visited at least once for $\beta$=1.}
  \label{fig:5}
\end{figure}

\begin{figure}
  \centering
  \includegraphics[width=1\textwidth]{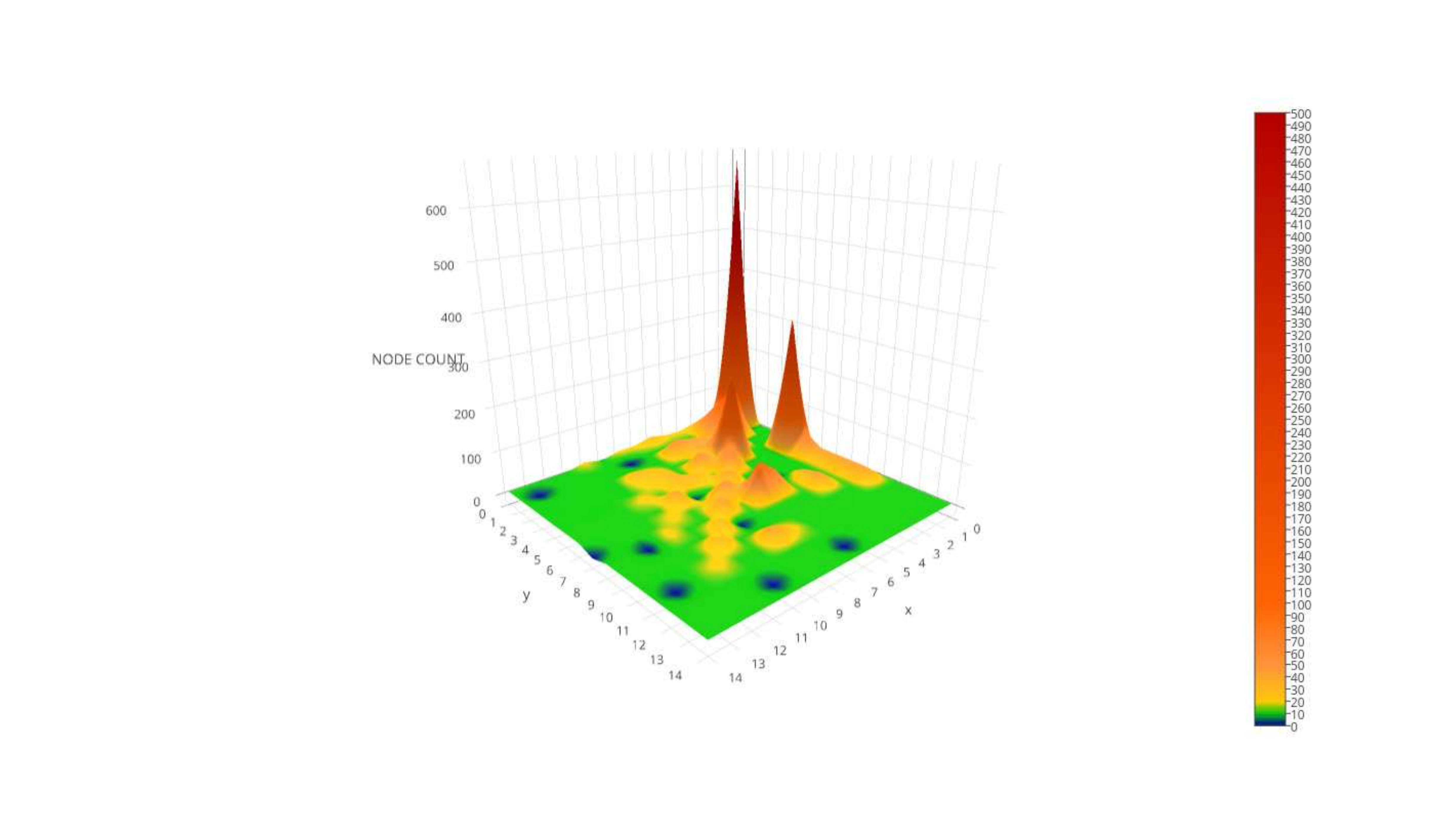}
  \caption{Number of visits for each node after they have all been visited at least 10 times for $\beta$=0.}
  \label{fig:6}
\end{figure}

\begin{figure}
  \centering
  \includegraphics[width=1\textwidth]{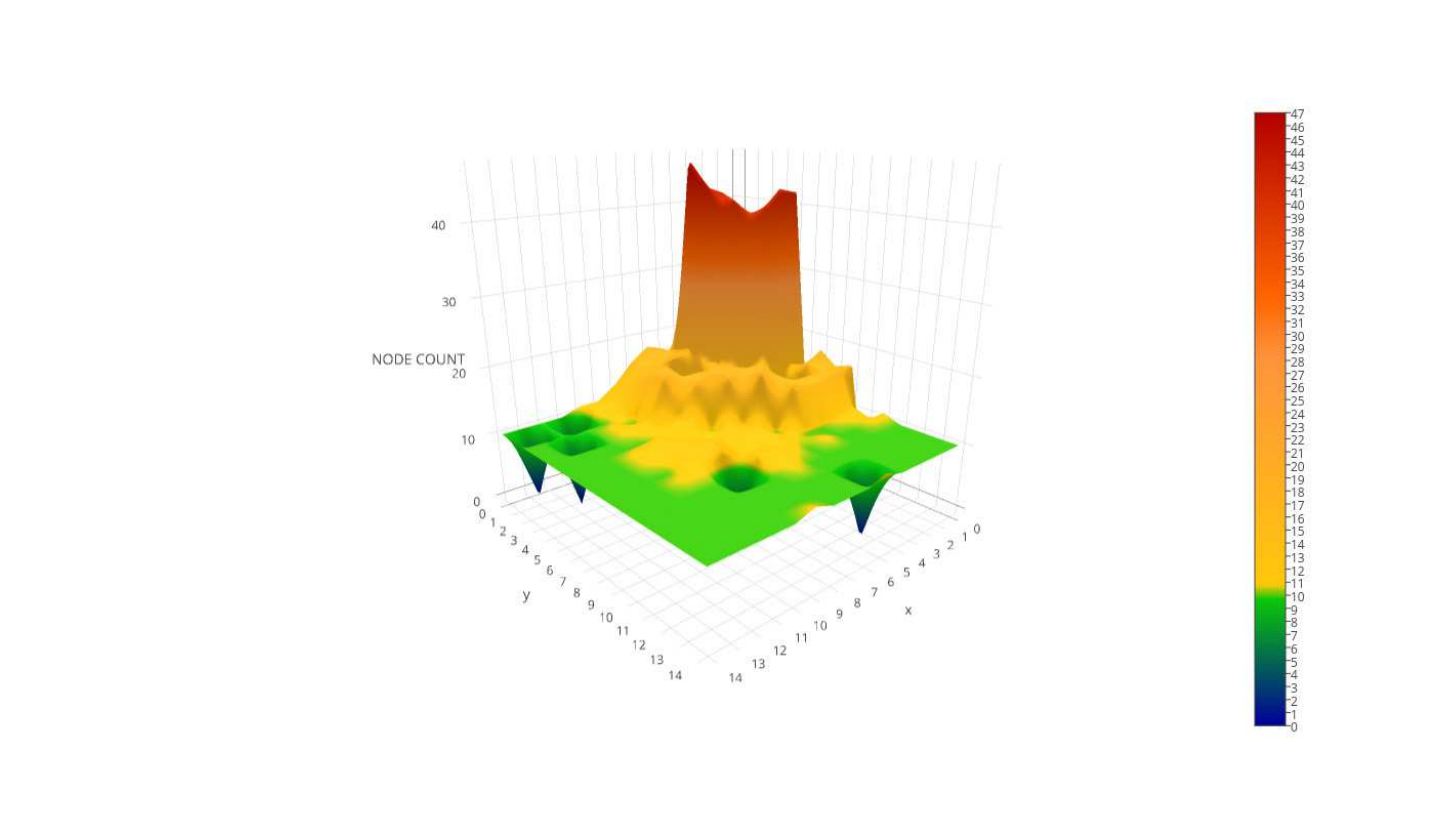}
  \caption{Total cost of the chain at each iteration until every node has been visited at least once for $\beta$=0.5.}
  \label{fig:7}
\end{figure}

\begin{figure}
  \centering
  \includegraphics[width=1\textwidth]{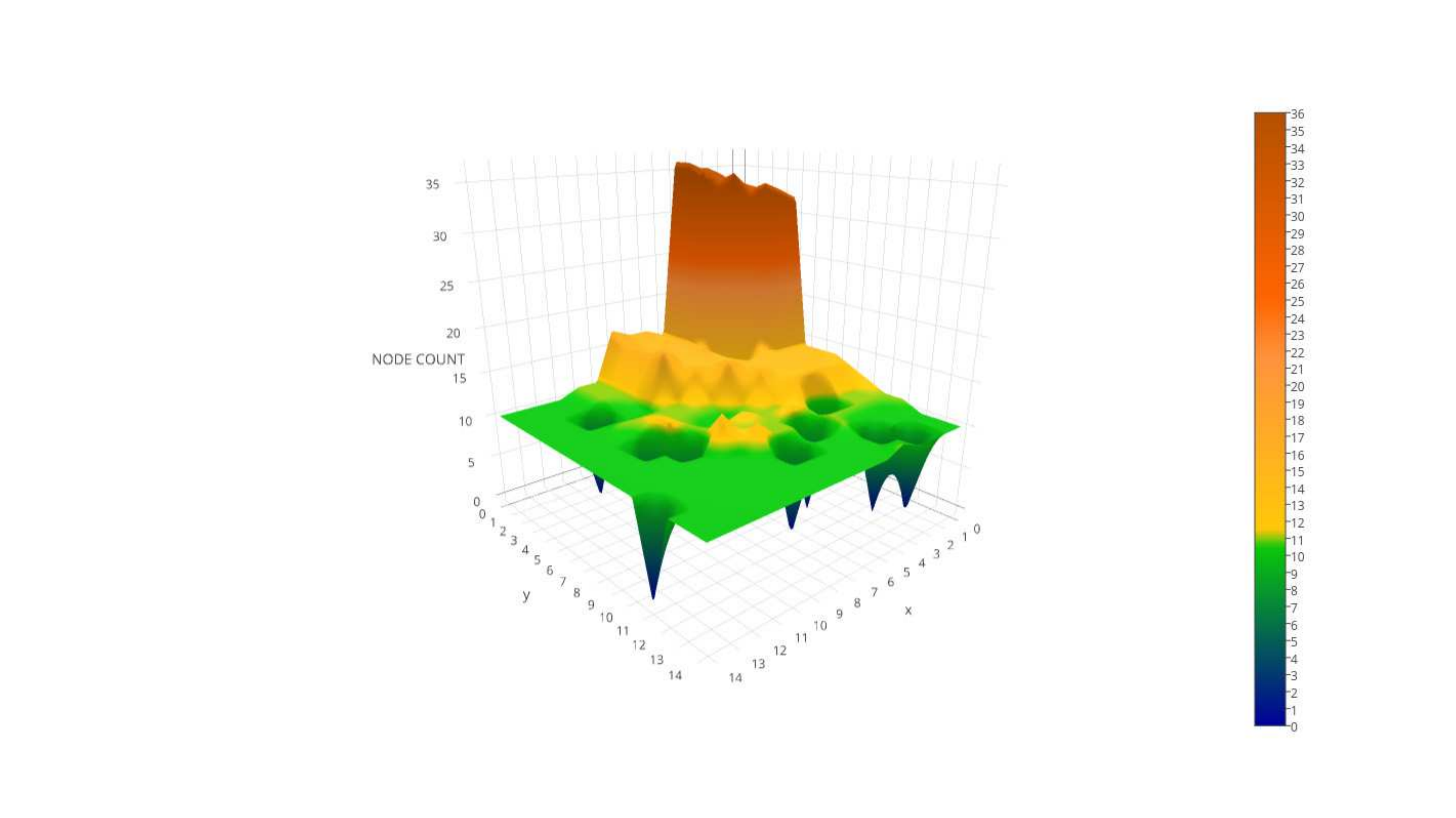}
  \caption{Total cost of the chain at each iteration until every node has been visited at least once for $\beta$=1.}
  \label{fig:8}
\end{figure}

\begin{figure}
  \centering
  \includegraphics[width=1\textwidth]{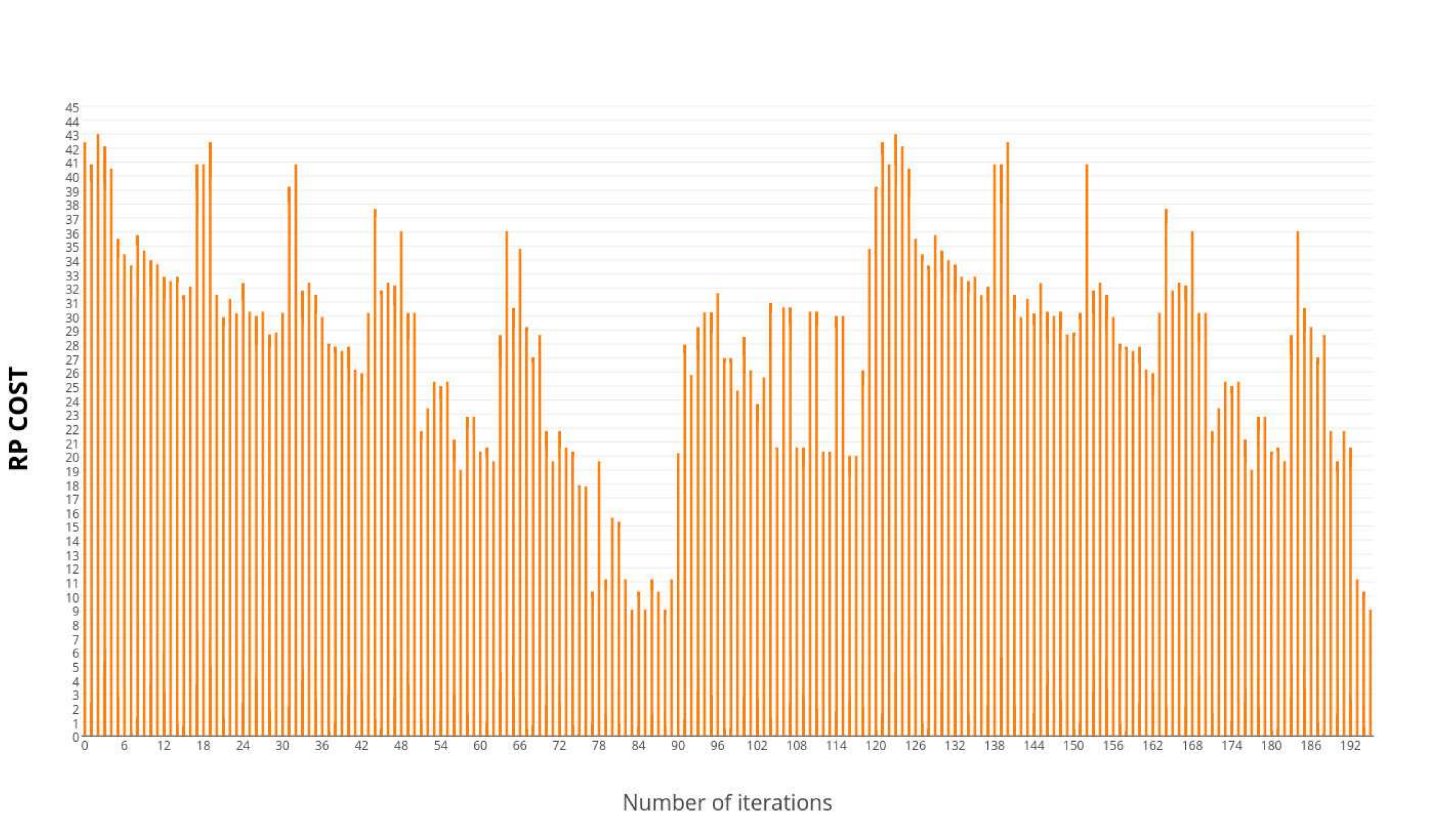}
  \caption{Total cost of the chain at each iteration until every node has been visited at least once for $\beta$=0.}
  \label{fig:9}
\end{figure}

\begin{figure}
  \centering
  \includegraphics[width=1\textwidth]{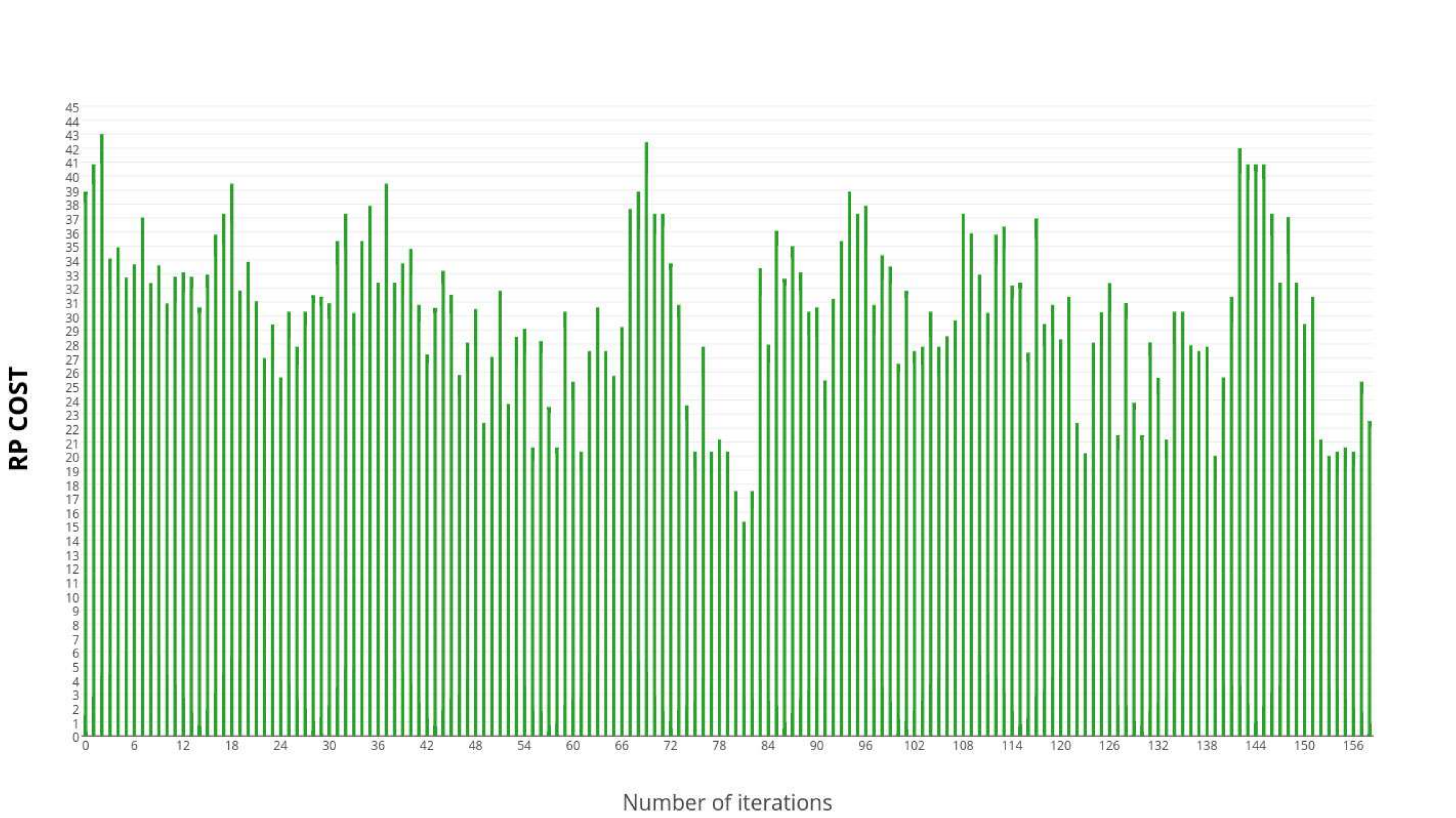}
  \caption{Number of visits for each node after they have all been visited at least once for $\beta$=0.5.}
  \label{fig:10}
\end{figure}

\begin{figure}
  \centering
  \includegraphics[width=1\textwidth]{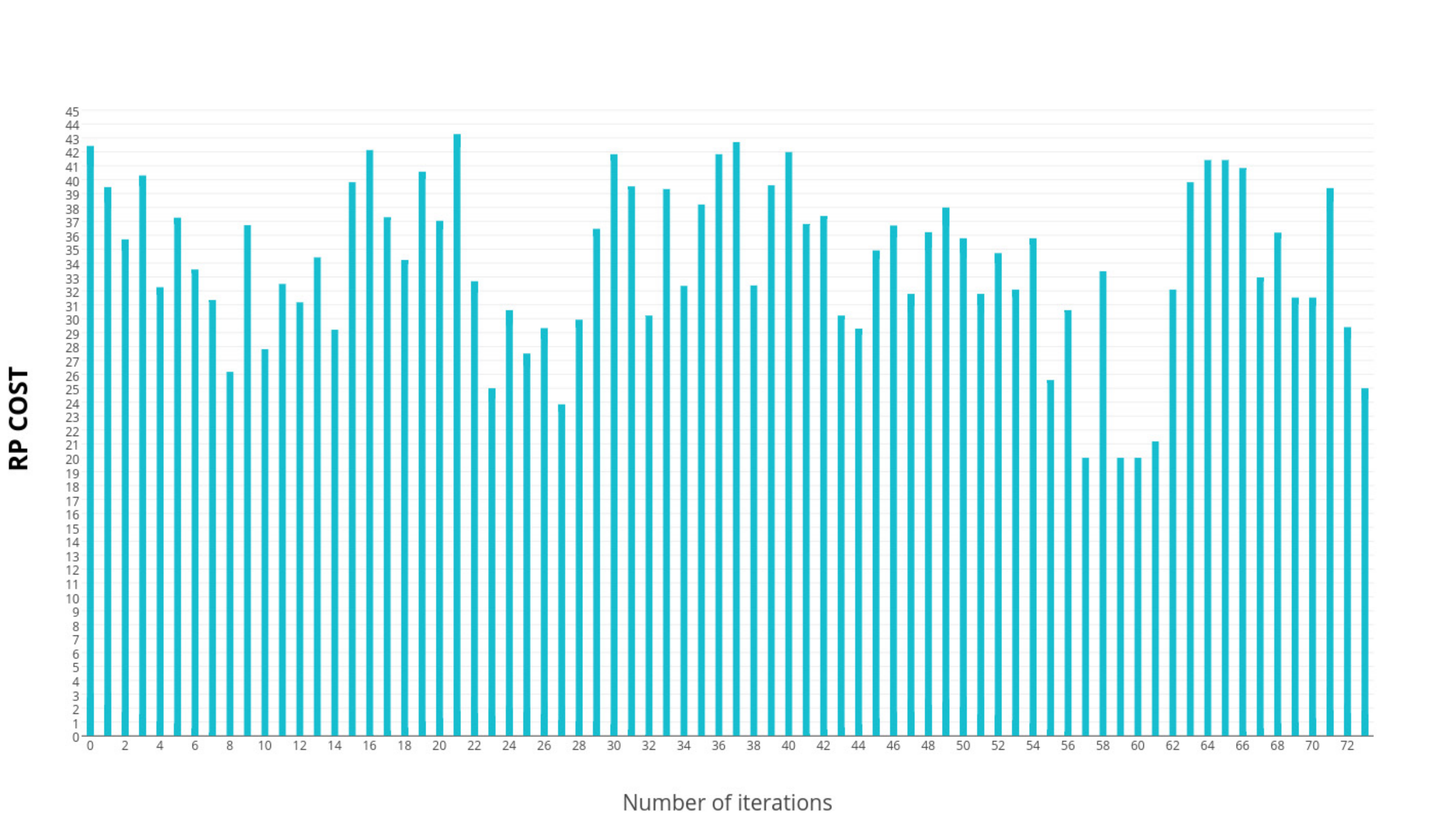}
  \caption{Number of visits for each node after they have all been visited at least once for $\beta$=1.}
  \label{fig:11}
\end{figure}

Let us focus on the impact of the parameter $\beta$.
Figures \ref{fig:3}, \ref{fig:4} and \ref{fig:5} represent the same environment with the same obstacle locations but different values for $\beta$. 
Nodes with a visit value of $0$ are either occupied by obstacles or unreachable because of obstacles themselves. 
We invite particular attention to the case in Figure \ref{fig:3} $n$ where $\beta = 0$, i.e., no coverage cost is taken into account and the problem reduces to the na\"{\i}ve case. 
In such a scenario, we observe large peaks of visits, especially in the nodes close to the base station and where the obstacles forms corridors. 
This is expected as it is always necessary to keep a valid relay chain, and these locations are fundamental to that purpose. 
For example, considering the case where the UAVs are tasked to visit all nodes at least once, the highest peaks happens exactly in those nodes and reach $77$ number of visits.
It took exactly $206$ iterations to reach the desired goal, and the average number of visits per node is $3.68$. 
As we increase the value of $\beta$ to 
$0.5$ (Figure \ref{fig:4}) and $1$ (Figure \ref{fig:5}), we notice that the peaky areas get smaller and so do the peak heights. 
In particular, the maximum visit value is lowered to $11$ and $4$
, respectively, while the number of iterations is reduced to $159$ and $74$. 
Furthermore, the average values are lowered to $3.14$ and $1.57$. 
Figures \ref{fig:6}, \ref{fig:7} and \ref{fig:8} report the results for the case in which each node must be visited at least $10$ times, again with the values of $\beta$ set to $0$, 
$0.5$, and $1$, respectively. 
In this case, we observe even larger peaks in these areas, but the same downward trend as $\beta$ increases, since the maximum number of visits goes from $673$ to 
$47$ and $36$. 
Back to the single visit case, let us consider the impact of $\beta$ at every iteration on the communication costs reported in Figures \ref{fig:9}, \ref{fig:10} and \ref{fig:11}. 
The mean and maximum costs are equal to $28.22$ and $43.00$ for $\beta = 0$,  
$30.22$ and $43.00$ for $\beta=0.5$, and 
$33.97$ and $43.28$ for $\beta = 1$. 
As expected, the average cost slowly rises when increasing $\beta$, but the worst case maximum communication cost remains steady. 
Comparing the results for $\beta = 0$ with $\beta = 1$ in the single visit case, we observe $64\%$ shorter time to reach the goal in terms of iterations, but only $17\%$ worse average communication cost. 
The worst-case maximum communication cost remains constant withing a reasonable margin of error. 
Similar results have been achieved on long-term operation with a higher target visits value.

\section{Conclusions}
\label{sec:conclusions}
In this Chapter we propose a novel approach to improve the performance of an UAVs swarm tasked to cover a given area in difficult contexts such as natural disasters. 
This is achieved by providing a relay chain to the base station so that UAVs do not need to rely on preexisting communication infrastructures. 
A number of supporting UAVs are employed to build the relay chain, but the proposed algorithm also allows them to provide assistance in the coverage task, decreasing the time required to completely cover the area without affecting too much the communication reliability. 
This is because the proposed algorithm does not seem to affect significantly the worst case communication cost of the chain, as highlighted by the reported experimental results. 
Moreover, the system allows UAVs to be momentarily busy with secondary tasks, such as target data acquisition, without necessarily leaving the swarm. 
On the robustness side, the architecture is robust to a missing master UAV, despite the fact that it presents a master-slave paradigm, and can also be extended to be more resilient to interruptions in the relay chain. 
Finally, the architecture allows for an easy and dynamic modification in the number of UAVs in the chain. 
This is useful both in the case of a scheduled task, as an UAV needing to recharge batteries, as well as in the case one UAV got separated from the swarm and needs to regroup. 

Future developments will deal with further increasing the robustness of the proposed architecture, by exploring in detail the best strategies to adopt when multiple UAVs are busy with secondary tasks or to avoid the chain to be interrupted. 
On the performance side, improved formulations of the coverage cost will be formalized. 
Testing in real-world is also in the works.